# Exemplar-Based Contrastive Self-Supervised Learning with Few-Shot Class Incremental Learning

Daniel T. Chang (张遵)

*IBM (Retired)* dtchang43@gmail.com

**Abstract:** Humans are capable of learning new concepts from only a few (labeled) exemplars, incrementally and continually. This happens within the context that we can differentiate among the exemplars, and between the exemplars and large amounts of other data (unlabeled and labeled). This suggests, in human learning, supervised learning of concepts based on exemplars takes place within the larger context of contrastive self-supervised learning (CSSL) based on unlabeled and labeled data. We discuss extending CSSL (1) to be based mainly on exemplars and only secondly on data augmentation, and (2) to apply to both unlabeled data (a large amount is available in general) and labeled data (a few exemplars can be obtained with valuable supervised knowledge). A major benefit of the extensions is that exemplar-based CSSL, with supervised finetuning, supports few-shot class incremental learning (CIL). Specifically, we discuss exemplar-based CSSL including: nearest-neighbor CSSL, neighborhood CSSL with supervised pretraining, and exemplar CSSL with supervised finetuning. We further discuss using exemplar-based CSSL to facilitate few-shot learning and, in particular, few-shot CIL.

## 1 Introduction

Humans are capable of learning new *concepts* from only a few *labeled exemplars*, generalizing without difficulty to new *unlabeled data*, incrementally and continually. This, however, does not take place in isolation. Rather, it occurs within the context that we can differentiate among the exemplars, and between the exemplars and large amounts of other data (unlabeled and labeled). In other words, in human learning, *supervised learning* of concepts based on exemplars takes place within the larger context of *contrastive self-supervised learning (CSSL)* based on unlabeled and labeled data.

Previously we discussed the use of *CSSL* for *concept (class) representation learning* [1-2]. (Note that our focus is on learning the *embodied level* of concept representations (i.e., *concept-oriented feature representations*) for *concrete concepts* (i.e., *classes*) [2]. As such, we use the terms '*concept*' and '*class*' interchangeably.) CSSL provides a promising approach to do so, since it: (1) uses data-driven associations, to get away from semantic labels, (2) supports incremental and continual learning, to get away from (large) fixed datasets, and (3) accommodates emergent objectives, to get away from fixed objectives (tasks). Traditional CSSL, however, has two major limitations when used for this purpose: (1) it is based solely on data augmentation, and (2) it is applied only to unlabeled data.

In this paper we discuss extending CSSL to remove these two limitations. First, CSSL is based mainly on *exemplars (labeled and unlabeled)* and only secondly on data augmentation. Second, CSSL is applied to both *unlabeled and labeled*

*data*. Doing so has two major benefits. First, exemplars represent similar instances within a semantic class which are crucial to learning concept-oriented feature representations that embed similar instances of the same class close to each other while separate embeddings from different classes. Further, the exemplars, being few, can be used to facilitate *class incremental learning (CIL)*. Second, it takes advantage of both unlabeled data (a large amount is available in general) as well as labeled data (a few exemplars can be obtained with valuable supervised knowledge).

We first discuss *exemplar-based CSSL* including: nearest-neighbor CSSL, neighborhood CSSL with supervised pretraining, and exemplar CSSL with supervised finetuning. We then discuss using exemplar-based CSSL to facilitate *few-shot learning* and, in particular, *few-shot CIL*.

## 2 Nearest-Neighbor (Exemplar) CSSL

CSSL constructs an objective for embeddings based on an assumed *semantic similarity* between positive pairs and *dissimilarity* between negative pairs, i.e., using instance discrimination. *InfoNCE loss* is commonly used as the contrastive loss in the instance discrimination setting. Specifically, the InfoNCE loss $\mathcal{L}_i^{InfoNCE}$ is defined as [3]:

$$\mathcal{L}_i^{InfoNCE} = -\log(\exp(z_i \cdot z_i^+/\tau) / (\exp(z_i \cdot z_i^+/\tau) + \sum_{z_i^- \in N_i} \exp(z_i \cdot z_i^-/\tau))),$$

where, for an embedded sample $z_i$, $(z_i, z_i^+)$ denotes a *positive pair* and $(z_i, z_i^-)$ denotes a *negative pair*. $N_i$ is the number of negative pairs. $z_i^+$ is drawn from the embedding distribution that represents the *semantically-similar* samples to $z_i$; $z_i^-$ is drawn from the embedding distribution that represents the *semantically-dissimilar* samples to $z_i$. $\tau$ is the softmax temperature. The overall loss is the average of all individual sample losses:

$$\mathcal{L}^{InfoNCE} = \frac{1}{M}\sum_{i=1}^{M} \mathcal{L}_i^{InfoNCE},$$

where M is the total number of samples. The goal is learning a representation that pulls positive pairs together in the embedding space, while separating negative pairs.

Most CSSL methods generate positive pairs using *data augmentation* applied to the same sample to obtain *multiple views* of the sample, which are assumed to be positive. SimCLR uses *two views* of the same sample as the positive pair. These two views are fed through an *encoder* to obtain the positive embedding pair $z_i$ and $z_i^+$. The negative pairs $(z_i, z_i^-)$ are formed using all the other embeddings. The infoNCE loss used in SimCLR, $\mathcal{L}_i^{SimCLR}$, is defined as [3]:

$$\mathcal{L}_i^{\text{SimCLR}} = -\log(\exp(z_i \cdot z_i^+/\tau) / \sum_{k=1}^{N} \exp(z_i \cdot z_k^+/\tau)),$$

where N is the number of all pairs. As before, the overall loss is the average of all individual sample losses. (Note that each embedding is l2 normalized before the dot product is computed in the loss.) Since the loss makes no assumption about the downstream tasks, the representations learned in principle can be used for various downstream tasks.

Data augmentations, however, cannot provide positive pairs for different viewpoints, deformations of the same sample, or *similar instances within a semantic class* [3], with the last limitation being the most critical from the perspective of *concept representation learning*. One needs to go beyond single instance positives to learn better features that are invariant to different viewpoints, deformations, and *intra-class variations*. This is done in *NNCLR*, which is discussed below.

## 2.1 NNCLR

*NNCLR (Nearest-Neighbor Contrastive Learning of visual Representations)* [3] samples the *nearest-neighbors* from the dataset in the embedding space, and treats them as positives. This provides more semantic variations than pre-defined data augmentations. Nearest-neighbors of a sample in the embedding space act as small *semantic perturbations* that are representative of *semantically-similar* samples (i.e., *class exemplars*) in the dataset.

To obtain nearest-neighbors, NNCLR utilizes a *support set Q* that keeps embeddings of a subset of the dataset in memory. The support set, which is implemented as a first-in-first-out queue, is used for nearest-neighbor search for retrieving cross-sample positives and it gets constantly replenished during training. The training process is shown below:

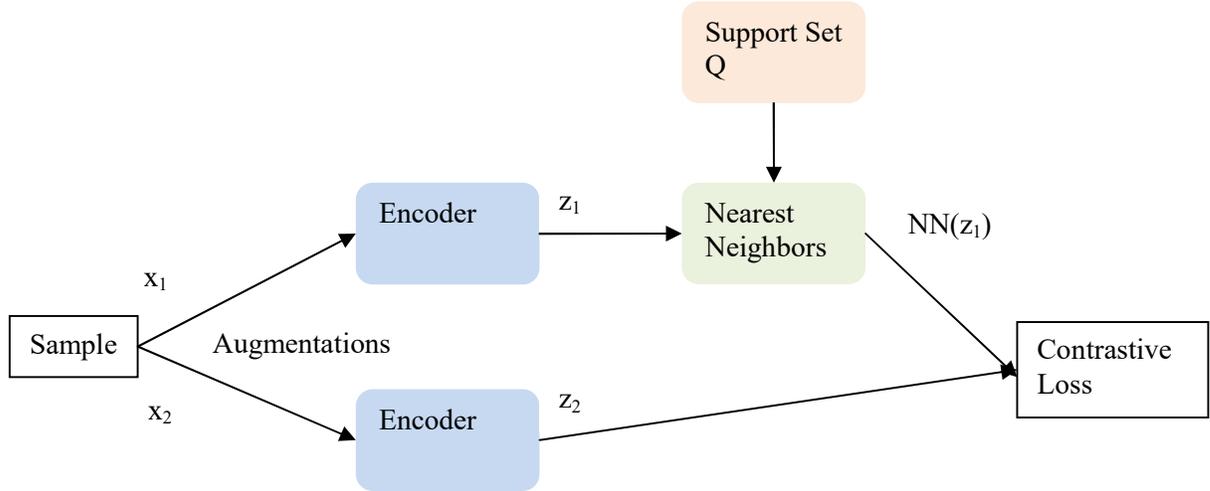

Building upon the SimCLR loss, the *NNCLR loss*, $\mathcal{L}_i^{NNCLR}$, is defined as [3]:

$$\mathcal{L}_i^{NNCLR} = -\log(\exp(NN(z_i, Q) \cdot z_i^+/\tau) / \sum_{k=1}^{N} \exp(NN(z_i, Q) \cdot z_k^+/\tau)),$$

where $NN(z, Q)$ is the *nearest-neighbor operator*:

$$NN(z, Q) = \arg\min(q \varepsilon Q) \|z - q\|_2.$$

As in SimCLR, each embedding is l2 normalized before the dot product is computed in the loss. Similarly, l2 normalization is applied before the nearest-neighbor operation. The overall loss is the average of all individual sample losses. In actual implementation, to boost the performance, $z_i^+$ is passed through a *prediction head g(.)* to produce embeddings $p_i^+ = g(z_i^+)$, and $p_i^+$, instead of $z_i^+$, is used in the loss.

Compared to SimCLR, NNCLR is much less dependent on complex augmentations since *nearest-neighbors* already provide richness in sample variations. For example, NNCLR with only crop augmentation performs as well as SimCLR with full augmentation. Thus, NNCLR reduces the reliance on data augmentation techniques drastically. For concept representation learning, NNCLR is most relevant in that sample variations are based on *semantic similarity*, i.e., *class exemplars*.

Note that we use both *unlabeled data* and *labeled data* for concept representation learning. Therefore, to be semantically consistent in treating data as *class exemplars*, we make a modification in using NNCLR. For *labeled samples* we (1) include samples with the same class labels as class exemplars and (2) exclude nearest-neighbors with a different class label as class exemplars. Therefore, for labeled samples, their class exemplars are samples with the *same class label* as well as *nearest-neighbors* which are unlabeled. (For *unlabeled samples*, their class exemplars are *nearest-neighbors*, with or without a class label.)

## 3 Neighborhood (Exemplar) CSSL with Supervised Pretraining

NNCLR (or specifically our modification of it) treats both labeled data and unlabeled data the same for the purpose of CSSL, except for selecting nearest-neighbors as class exemplars. However, it is desirable to *leverage the labeled data* (collected from *known classes*) to explore the unlabeled data for discovery of *novel classes*. This can be achieved by *pretraining a feature extractor with supervision* using the labeled data and then applying *CSSL* on both the labeled and unlabeled data. This is done in *NCL,* which is discussed below.

### 3.1 NCL

*NCL (Neighborhood Contrastive Learning)* [4] addresses *Novel Class Discovery (NCD)*, the task of inferring *novel classes* in an *unlabeled dataset* by leveraging from prior knowledge of a *labeled dataset* containing different *known classes*. NCD assumes the availability of two datasets: a *labeled dataset* $D^l$ containing known classes $C^l$ and an *unlabeled dataset* $D^u$ containing novel classes $C^u$. The sets of classes $C^l$ and $C^u$ are disjoint. The goal of NCD is to learn (unsupervised) the $C^u$ class representation from $D^u$, leveraging the $C^l$ class representation learned (supervised) from $D^l$.

NCL proposes a holistic learning framework that uses *contrastive loss* formulation to learn *discriminative features* from both labeled and unlabeled data. It enforces a sample to be close to its *correlated view* (augmented-positive) and its *nearest-neighbors* (pseudo-positives), as well as to be far from the negatives. (Note that for both labeled and unlabeled samples, *two correlated views* are generated for each sample using data augmentation.) There are two key ideas. The first idea is that the *nearest-neighbors* of a sample in the embedding space most likely belong to the *same semantic class* of the sample (i.e., *class exemplars*), and can be considered as positives. (Note that this is also a key idea of *NNCLR*.) The second idea is that *better selection of negatives* can improve the performance of contrastive learning. Since *labeled samples* of the known classes are a.k.a. true negatives of any unlabeled sample, they are treated as *hard negatives*.

For clarity and ease of understanding, we break down the *NCL framework* into *three stages*, with a shared *feature extractor (encoder)*. In the *first (pretraining) stage*, high-level features are learned through *supervised classification* using the labeled dataset. The network is optimized using the standard *cross-entropy (CE) loss* and the *consistency (CS) loss* [4]. The CS loss enforces the network produce similar predictions for a sample $x_i$ and its correlated view $\hat{x}_i$. The first stage is shown below:

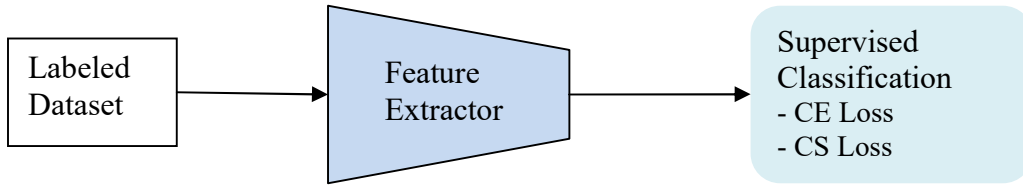

Note that if the labeled dataset is *low volume*, there is a major issue of *overfitting*. For discussion, see 5 Few-Shot Learning with Exemplar CSSL.

In the *second (contrastive learning) stage*, contrastive learning is applied to both the unlabeled dataset and the labeled dataset. First, *neighborhood contrastive learning (NCL)* is applied to the unlabeled dataset. The *NCL loss* [4] consists of two terms. The first term is the contrastive loss for the *correlated view* (augmented-positive); the second term is the contrastive loss for the *nearest-neighbors* (pseudo-positives). Second, *"supervised" contrastive learning (SCL)* is applied to the labeled dataset. The *SCL loss* [4] is the contrastive loss for the positives that have the *same class label* as the labeled sample. The second stage is shown below:

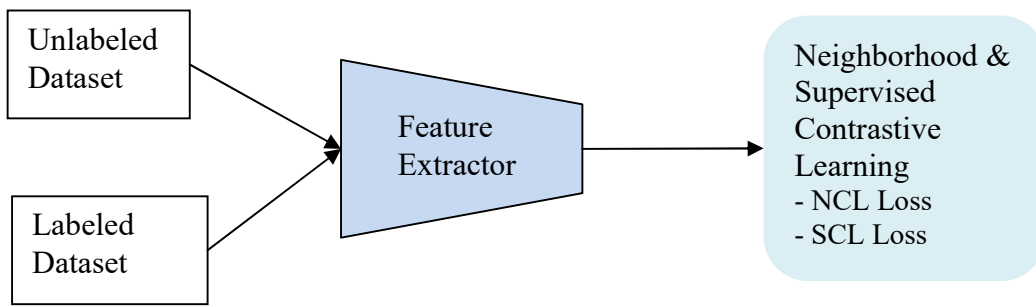

Finally, in the *third (NCD) stage*, the cosine similarity of the features is used to estimate the pairwise pseudo-label, which is then compared to the inner product of the outputs of the unlabeled head to compute the *binary cross-entropy (BCE) loss* [4]. The network is optimized using the BCE loss and the *consistency (CS) loss* [4]. As in the first stage, the CS loss enforces the network produce similar predictions for a sample $x_i$ and its correlated view $\hat{x}_i$. The third stage is shown below:

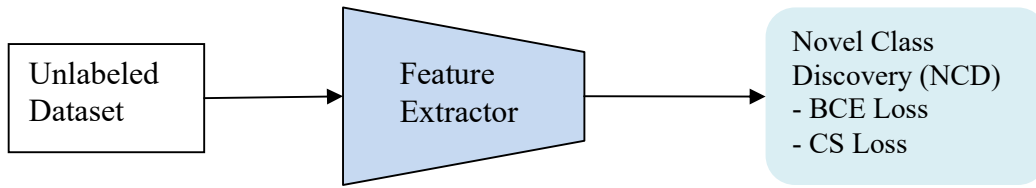

## 4 Exemplar CSSL with Supervised Finetuning

To *leverage the labeled data* (collected from *known classes*) to explore the unlabeled data for discovery of *novel classes*, instead of pretraining a feature extractor with supervision using the labeled data (as discussed in the previous section), an alternative is to apply *CSSL* on both the labeled and unlabeled data first and then *finetune* the learned representation using the *labeled data with supervision*. This is found to be a good paradigm for learning from *few labeled exemplars* while making best use of *a large amount of unlabeled data* [5]. It also avoids the *overfitting* problem associated with (pre)training low-volume labeled data, mentioned in the previous section.

In general, representations learned through CSSL enable fast finetuning to multiple downstream tasks, and lead to better generalization and calibration [6]. Our interest is in: (1) applying *exemplar CSSL* (e.g., *NNCLR*) to both the labeled dataset (containing data on known classes) and the unlabeled dataset (containing data on novel classes) to learn *task-agnostic* representations, and (2) *finetuning* the learned representations through supervised learning on the labeled dataset. The finetuned representations remain *task-agnostic*, though they are *concept (class) oriented* since exemplar CSSL embeds similar instances of the same class close to each other while separate embeddings from different classes.

The learning process consists of *two stages*, with a shared *feature extractor (encoder)*. The *first (CSSL) stage* is shown below:

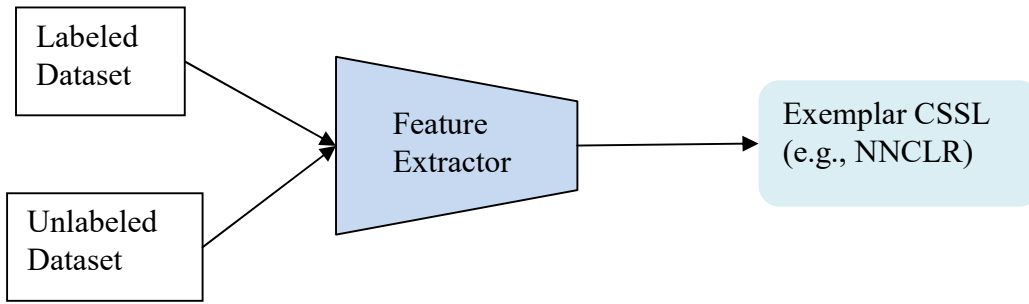

and the *second (supervised finetuning)* stage:

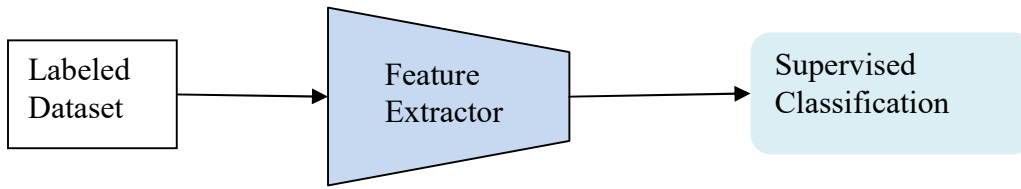

A major advantage of the two-stage learning process of *exemplar CSSL with supervised finetuning* is that it naturally supports *few-shot learning* and, in particular, *few-shot CIL*. This is discussed in the next two sections.

## 5 Few-Shot Learning with Exemplar CSSL

For *labeled data*, as mentioned in the beginning, our main interest is in use cases where the labeled data represent *class exemplars* that are *few* (so that they can be obtained without too much difficulty). As such, only *low-volume* labeled datasets are available for use in *supervised pretraining* or *finetuning* discussed in the previous two sections.

*Few-shot learning* [7-8] aims for optimization methods and models that can learn efficiently to recognize patterns in the low data regime. The main issue with few-shot learning is *overfitting* to training data, resulting in *poor accuracy* on test data. To overcome data scarcity, *transfer learning* is used by all few-shot learning methods, which uses large-scale training data on some *base classes* and then adapt to *novel classes* with few training data. A common way to do this is using *two learning stages*, with a shared *feature extractor*: in the first (*base*) learning stage, train a network on the base class data, and in the second (*few-shot*) learning stage, extend the network on the novel class data. Both stages are usually done using *supervised learning* on labeled data.

An approach for improving few-shot learning with CSSL is proposed [8] by exploiting the complementarity of few-shot learning and CSSL. It uses *CSSL as an auxiliary task in a few-shot learning pipeline*, enabling feature extractors to learn richer and more transferable representations while still using few labeled exemplars. This is done by adding a *self-supervised loss* to the training loss during its first learning stage. Moreover, one can include extra *unlabeled (base class) data* to the first learning stage. At the extreme, one can even use only unlabeled (base class) data in the first learning stage, thus removing the use of labeled base class data altogether. The first (*base*) learning stage is shown below:

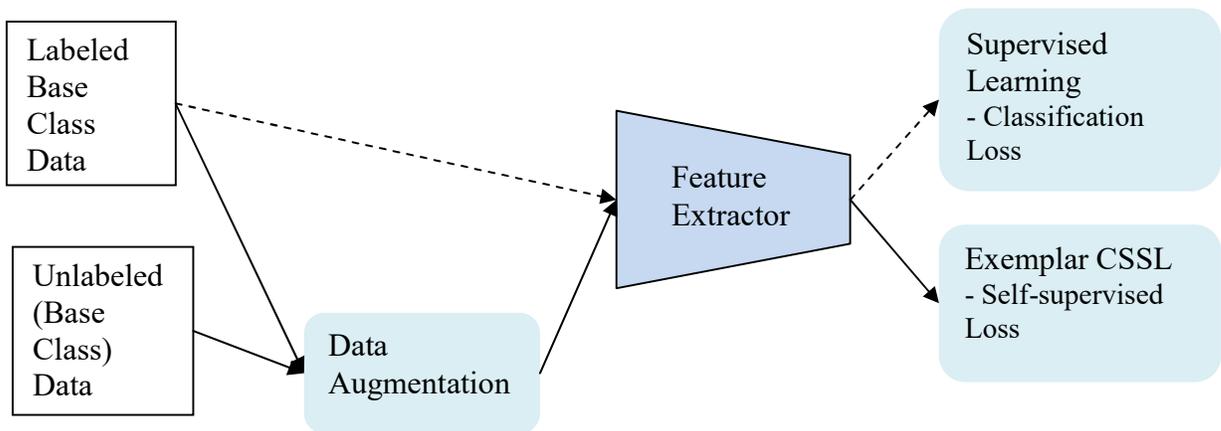

The second (*few-shot*) learning stage extends the feature extractor on the novel class data, as shown below:

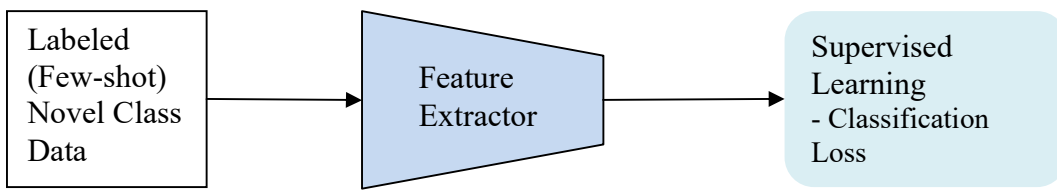

Note that we use *exemplar CSSL*, not just CSSL as in [8], in the first (base) learning stage. Exemplar CSSL, being based mainly on class exemplars (labeled or unlabeled) and only secondly on data augmentation, matches well with supervised base class learning, both being *class centric*. Note also that if we use only exemplar CSSL, without supervised base class learning, the learning process is identical to that discussed in 4 Exemplar CSSL with Supervised Finetuning. (In which case, in the diagram for the first (base) learning stage, the dashed lines and the "Supervised Learning" box are removed.)

# 6 Few-Shot CIL

In *CIL* [1], a model learns tasks continually, with each task containing a batch of *new classes*. In *few-shot CIL*, only the training set of the first task (a.k.a., the *base task*) may have large-scale training data, while other subsequent tasks (a.k.a., the *new task*) just contains *a few class exemplars*. Few-shot CIL requires transferring knowledge from old classes to new classes in solving the *catastrophic forgetting* problem. The requirement is generic to CIL. The additional challenge of few-shot learning is the *overfitting* problem mentioned in the previous section.

The *two-stage learning process*, discussed in the previous two sections, naturally supports few-shot CIL. The first (*CSSL / base*) stage supports few-shot learning by using *large-scale base-class (labeled or unlabeled)* training to prevent *overfitting* few-shot novel class training data in the second stage. It is used once. The second (*supervised finetuning / few-shot*) stage can be used repeatedly for (few-shot) CIL. The *catastrophic forgetting* problem, with base classes and old novel classes, can be mitigated using, among others, the knowledge distillation approach.

Various *knowledge distillation* techniques have been developed for *few-shot CIL* to transfer knowledge from base classes and old novel classes to new novel classes, thus allowing the use of only *new novel class data* to train the second stage, incrementally and continually. In the following, we discuss two recent approaches to knowledge distillation for few-shot CIL. Note that they are *class (neighborhood) centric* and thus well-suited for use with *exemplar CSSL*.

## 6.1 TOPIC

The *TOPIC (TOpology-Preserving knowledge InCrementer)* framework [9] is inspired by discoveries in *cognitive neural science* which reveal the importance of *topology preservation* for maintaining the memory of the old knowledge and thus avoiding catastrophic forgetting. TOPIC uses a *neural gas (NG)* network for feature representation, which can learn and preserve the *topology of the feature space* formed by different classes. It mitigates the forgetting of the old classes by stabilizing NG's topology, and it improves the representation learning for *few-shot novel classes* by growing and adapting NG to new training samples.

*NG* defines an undirected graph. Each *vertex* $v_i$ is assigned a centroid vector describing the location of $v_i$ in feature space. The *edge set E* stores the neighborhood relations of the vertices. If $v_i$ and $v_j$ are topologically adjacent, $e_{ij} = 1$; otherwise, $e_{ij} =$

0. Initially, NG is learnt for *base classes*. It preserves the topology of feature space by competitive Hebbian learning. Then NG incrementally grows for *new novel classes* by inserting new nodes and edges. To maintain the topology, the shift of NG nodes is penalized by an *anchor-loss term*. To address the overfitting problem, a *min-max loss term* is introduced to constrain the feature vector of a new sample and the centroid vectors of $v_i$ and its neighbors.

### 6.2 ERDIL

The *exemplar relation distillation incremental learning (ERDIL)* framework [10] is based on the observation that *relational information between exemplars* provides an effective way to constrain not only the locations of exemplars in the feature space but also their relations. It uses *relational knowledge distillation* to balance the tasks of old-knowledge preserving and new-knowledge adaptation. First, it constructs an *exemplar relation graph (ERG)* to represent the knowledge learned by the network and gradually updates the relation graph for novel classes learning. Second, an *exemplar relation loss function* for discovering the relation knowledge between different classes is introduced to learn and transfer the structural information in relation graph.

*ERG* is made up of *vertices* of feature vectors of selected class exemplars and *edges* linking these vertices. The edges of ERG are *weighted by the angles* between the linked exemplar feature vectors. A *nearest-neighbor-based exemplars selection mechanism* is used to construct the directed ERG such that each class only contains the *K most representative exemplars* in the ERG.

The structural knowledge in ERG can be divided into two categories: absolute relations and relative relations. The *absolute relation* is the distance formed by the double vertices in ERG and the *relative relation* is the angle formed by the triplet vertices (a pair of edges) in ERG. Relative relations are more flexible because when the angle relationship remains unchanged, the overall scaling of the distance will not destroy the topology of the feature space. ERDIL, therefore, explores *relative relations in ERG* for knowledge distillation by defining an *exemplar relation loss* function [10] based on relative relations to learn and transfer the structural information in ERG.

### 7 Conclusion

Exemplar CSSL (e.g., NNCLR) provides an excellent approach to learn concept representations at the embodied level (i.e., concept-oriented feature representations). Exemplars (labeled or unlabeled) represent similar instances within a concept (class) and, therefore, facilitate embedding similar instances of the same class close to each other while separating embeddings from different classes.

The two-stage learning process of exemplar CSSL with supervised finetuning naturally supports few-shot learning and, in particular, few-shot CIL. The first (CSSL) stage, with large-scale base class training, supports few-shot learning by preventing overfitting in the second stage; the second (supervised finetuning) stage, with few-shot novel class training incrementally and continually, supports CIL by using, among others, the knowledge distillation approach to avoid catastrophic forgetting.

**Acknowledgement:** Thanks to my wife Hedy (郑期芳) for her support.